\documentclass[lettersize,journal]{IEEEtran}
\usepackage{amsmath,amsfonts}
\usepackage{algorithmic}
\usepackage{algorithm}
\usepackage{array}
\usepackage[caption=false,font=normalsize,labelfont=sf,textfont=sf]{subfig}
\usepackage{textcomp}
\usepackage{stfloats}
\usepackage{url}
\usepackage{verbatim}
\usepackage{graphicx}
\usepackage{cite}
\usepackage{algorithm}
\usepackage{algorithmic}

\usepackage{colortbl}  
\usepackage{xcolor}
\usepackage{array}

\usepackage{graphicx}
\usepackage{multirow}
\usepackage{bbding}
\usepackage{booktabs} 
\usepackage{xcolor}
\newcommand{\eg}{\textit{e}.\textit{g}.}
\newcommand{\etal}{\textit{et al}.}
\newcommand{\ie}{\textit{i}.\textit{e}.}

\newcommand{\aka}{\textit{a.k.a.}}
\definecolor{hollywoodcerise}{rgb}{0.96, 0.0, 0.63}
\definecolor{lasallegreen}{rgb}{0.03, 0.47, 0.19}
\definecolor{hanpurple}{rgb}{0.32, 0.09, 0.98}
\definecolor{green(pigment)}{rgb}{0.0, 0.65, 0.31}
\usepackage[pagebackref=true,breaklinks=true,letterpaper=true,colorlinks,bookmarks=false]{hyperref}
\hypersetup{colorlinks,linkcolor={red},citecolor={hanpurple},urlcolor={red}}  
\begin{document}

\title{MAGIC++: Efficient and Resilient Modality-Agnostic Semantic Segmentation via Hierarchical Modality Selection}
% Centering the Value of Every Modality at Every Granularity: Towards Efficient and Resilient Modality-agnostic Semantic Segmentation with Hierarchical Modality Selection}

\author{Xu Zheng$^{1}$, Yuanhuiyi Lyu$^{1}$, Lutao Jiang$^{1}$, Jiazhou Zhou$^{1}$, Lin Wang$^{3}$$^\dagger$, Xuming Hu$^{1,2}$$^\dagger$
%\thanks{This paper was produced by the IEEE Publication Technology Group. They are in Piscataway, NJ.}% <-this % stops a space
\\
$^{1}$AI Thrust, HKUST(GZ)  \quad $^{2}$Dept. of CSE, HKUST \quad $^{3}$Dept. of EEE, NTU\\
\thanks{$^\dagger$: Corresponding Author}
\thanks{A preliminary version of this work has appeared in ECCV 2024~\cite{zheng2025centering}.}
% \thanks{Manuscript created October, 2020; This work was developed by the IEEE Publication Technology Department. This work is distributed under the \LaTeX \ Project Public License (LPPL) ( http://www.latex-project.org/ ) version 1.3. A copy of the LPPL, version 1.3, is included in the base \LaTeX \ documentation of all distributions of \LaTeX \ released 2003/12/01 or later. The opinions expressed here are entirely that of the author. No warranty is expressed or implied. User assumes all risk.}
% \thanks{Xu Zheng is with the AI Thrust, HKUST(GZ), Guangdong, China. E-mail: zhengxu128@gmail.com. 
% % Lin Wang is with AI/CMA Thrust, HKUST(GZ) and Dept. of CSE, HKUST, Hong Kong SAR, China, E-mail: linwang@ust.hk.
% }
}

\markboth{Journal of \LaTeX\ Class Files,~Vol.~14, No.~8, August~2021}%
{Shell \MakeLowercase{\textit{et al.}}: A Sample Article Using IEEEtran.cls for IEEE Journals}


\maketitle

\begin{abstract}
In this paper, we address the challenging modality-agnostic semantic segmentation (MaSS), aiming at centering the value of every modality at every feature granularity. 
Training with all available visual modalities and effectively fusing an arbitrary combination of them is essential for robust multi-modal fusion in semantic segmentation, especially in real-world scenarios, yet remains less explored to date. 
Existing approaches often place RGB at the center, treating other modalities as secondary, resulting in an asymmetric architecture. However, RGB alone can be limiting in scenarios like nighttime, where modalities such as event data excel. Therefore, a \textbf{resilient} fusion model must dynamically adapt to each modality's strengths while compensating for weaker inputs.
To this end, we introduce the \textbf{MAGIC++} framework, which comprises two key \textbf{plug-and-play} modules for effective multi-modal fusion and hierarchical modality selection that can be equipped with various backbone models. 
Firstly, we introduce a multi-modal interaction module to efficiently process features from the input multi-modal batches and extract complementary scene information with channel-wise and spatial-wise guidance. On top, a unified multi-scale arbitrary-modal selection module is proposed to utilize the aggregated features as the benchmark to rank the multi-modal features based on the similarity scores at hierarchical feature spaces. This way, our method can eliminate the dependence on RGB modality at every feature granularity and better overcome sensor failures and environmental noises while ensuring the segmentation performance. 
Under the common multi-modal setting, our method achieves state-of-the-art performance on both real-world and synthetic benchmarks. Moreover, our method is superior in the novel modality-agnostic setting, where it outperforms prior arts by a large margin, \ie, +2.19\% on MUSES and +7.25\% on DELIVER.
\end{abstract}


\section{Introduction}
Nature has demonstrated that diverse sensory and visual processing capabilities are crucial for understanding complex environments~\cite{duan2022multimodal, su2023recent,zhou2024eventbind}. Accordingly, intelligent systems like robots or autonomous vehicles require multi-sensor setups, including RGB, LiDAR, and event cameras, to achieve robust scene perception and understanding, particularly for the dense pixel-wise semantic segmentation tasks~\cite{wang2023multi, alonso2019ev, jia2023event,zhu2024customize,zheng2024learning}. Every specific sensor provides unique characteristics and advantages, which complement each other in challenging scenarios, such as low light conditions in nighttime and fast motions~\cite{wang2021survey, zheng2023deep}.

\begin{figure}[t!]
    \centering
    \includegraphics[width=\linewidth]{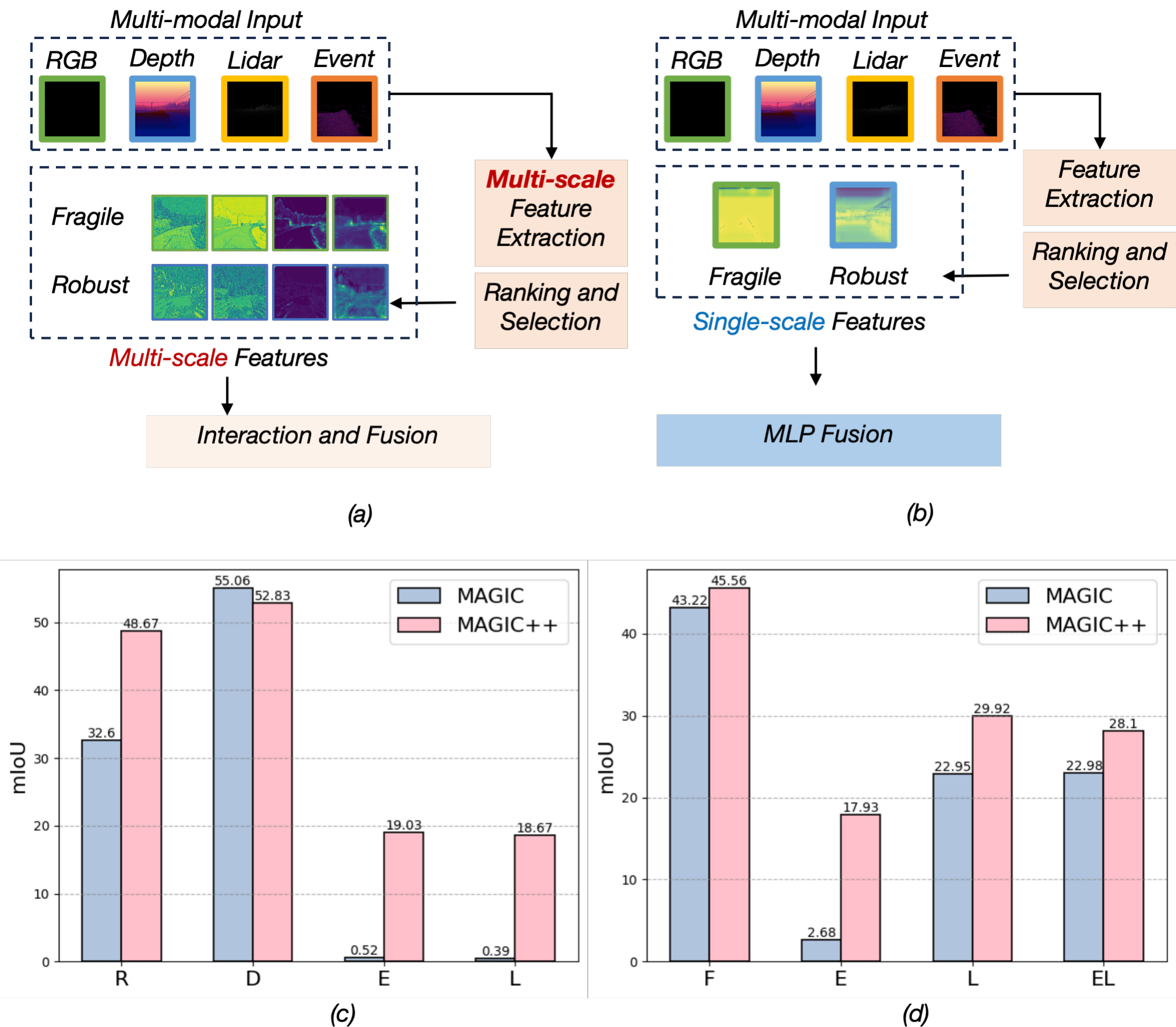
    }
    \caption{(a) MAGIC++ framework with multi-scale arbitrary modality selection and multi-modal interaction modules; 
    (b) MAGIC framework with single scale arbitrary modality selection and MLP based multi-modal fusion;  
    (b) \& (c): Performance comparison between MAGIC and MAGIC++ frameworks on DELIVER~\cite{zhang2023delivering} and MUSES~\cite{brodermann2025muses} datasets. 
    }
    \label{cover_fig}
\end{figure}

Initial attempt in achieving multi-modal fusion focus on designing tailored fusion architectures for specific sensor pairs, such as RGB-depth~\cite{song2022improving}, RGB-Lidar~\cite{li2023mseg3d}, RGB-event~\cite{zhang2021issafe}, and RGB-thermal~\cite{hui2023bridging}. 
While effective for these fixed combinations, these approaches often lack flexibility and scalability when incorporating additional sensors.
Given the demand for versatile multi-modal systems, enabling the fusion of arbitrary sensor combinations is increasingly valuable for robust multi-modal segmentation. However, this research area remains under-explored. Only recently have a few works attempt to address this challenge by positioning the RGB modality as primary, with others treated as auxiliary inputs~\cite{zhu2023visual, zhang2022cmx, zhang2023delivering}. This design naturally results in an RGB-centric framework, typically with a unified RGB-X pipeline and either a distributed or asymmetric two-branch structure.
A representative approach, CMNeXt~\cite{zhang2023delivering}, introduces a self-query hub to selectively extract relevant information from auxiliary modalities, which is then fused with the primary RGB input for enhanced segmentation performance. 

However, the RGB modality can under-perform in certain conditions, such as nighttime, as shown by the visualized features in Fig.~\ref{cover_fig} (a). In contrast, alternative sensors provide distinct advantages that improve scene understanding in challenging settings. For instance, depth cameras are reliable in low-light conditions and deliver spatial information unaffected by ambient lighting, making them particularly useful for nighttime applications.
This highlights that \textit{\textbf{fully recognizing the value of each modality}} is essential for leveraging their combined strengths to achieve modality-agnostic segmentation. Thus, it becomes crucial for the fusion model to identify and utilize both the robust and fragile modalities at every feature granularity to construct a more resilient multi-modal framework. The robust features contribute to enhancing segmentation accuracy, while the fragile features help reinforce the framework’s resilience against missing modalities.

To address these challenges, we propose an efficient and resilient \textbf{M}odality-\textbf{ag}nost\textbf{ic} (\textbf{MAGIC++}) segmentation framework that is compatible with a wide range of backbone models, \eg, SegFormer, Swin Transformer, and Pyramid Vision Transformer (PVT), spanning from lightweight to high-performance architectures. Our approach incorporates two plug-and-play modules designed to enhance multi-modal learning and bolster modality-agnostic robustness in segmentation models. 
First, we introduce the Multi-modal Interaction Module (MIM), which efficiently integrates features from multiple modalities through channel-wise and spatial-wise matching. This module extracts complementary scene information without relying on any specific modality, ensuring flexibility and adaptability across diverse scenarios. 

\begin{figure*}[t!]
    \centering
    \includegraphics[width=\textwidth]{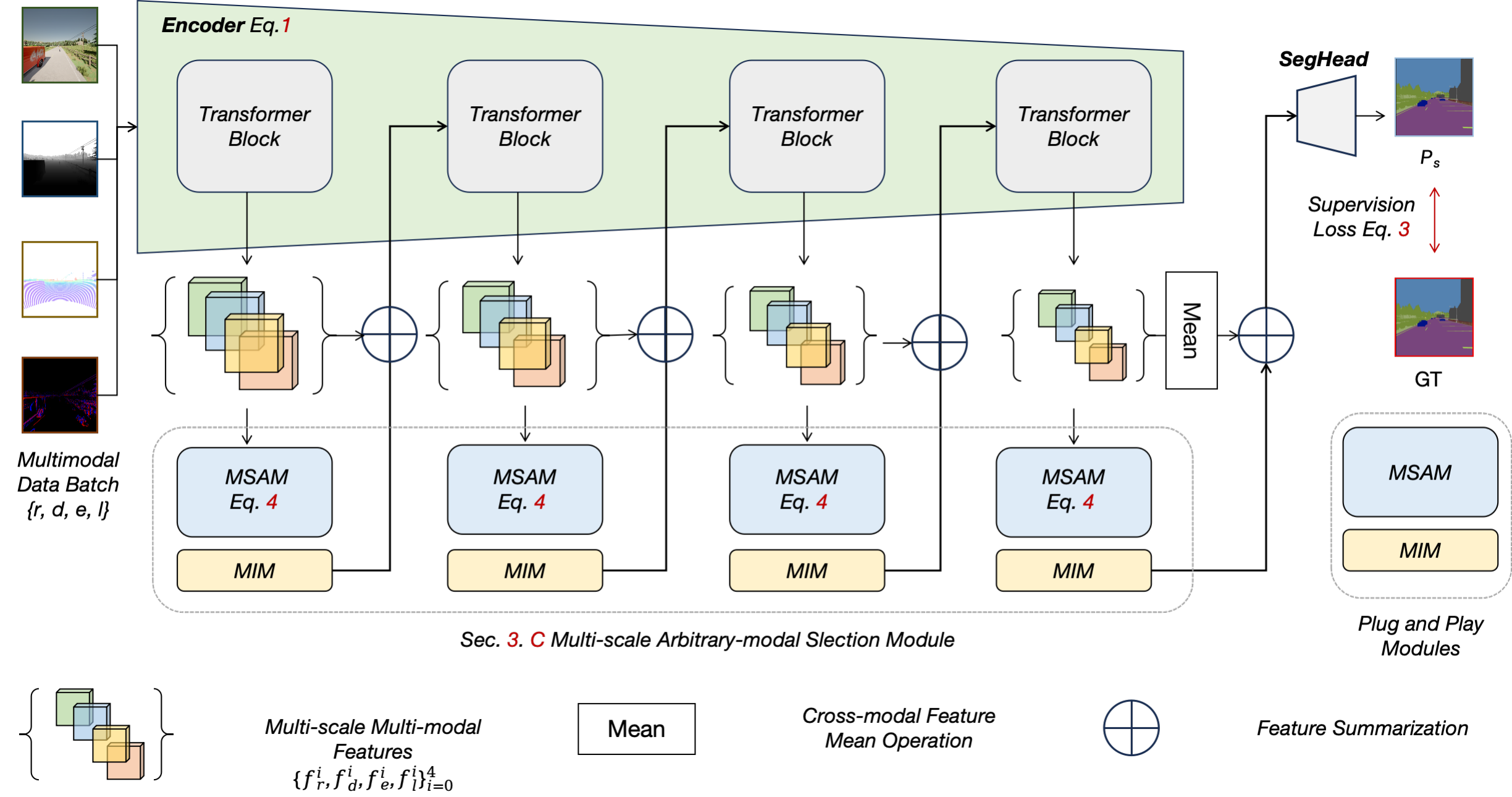}
    \caption{Overall framework of MAGIC++ framework, incorporates plug-and-play multi-modal interaction module (MIM) and multi-scale arbitrary-modal selection module (MASM).
    }
    \label{fig:overall}
\end{figure*}

Building upon MIM, we present the Multi-scale Arbitrary-modal Selection Module (MASM), which dynamically fuses features across multiple granularities during training to enhance the backbone model's robustness to arbitrary-modal input at inference. MASM utilizes the integrated features from MIM as a reference to rank multi-modal features based on similarity scores within hierarchical feature spaces, \eg, the four scale features in SegFormer's backbone model as shown in Fig.~\ref{fig:overall}. It then merges the top-ranked (most robust) and last-ranked (most fragile) features to generate predictions. This process enables the fusion model to effectively differentiate between \textbf{robust} and \textbf{fragile} modalities. Incorporating both robust and fragile modalities allows the model to learn a more \textbf{resilient} multi-modal framework, where robust features improve segmentation accuracy, and fragile features enhance the framework’s resilience against missing modalities. This modality-agnostic design reduces reliance on RGB inputs and mitigates the effects of sensor failures, as illustrated in Fig.~\ref{cover_fig} (b) and (c).
Additionally, MASM incorporates MIM's predictions with ground truth to soften the supervision for its own outputs, ensuring stable training convergence.

We conduct extensive experiments on both synthetic and real-world benchmarks~\cite{zhang2023delivering,brodermann2025muses}, including RGB, Depth, LiDAR, and event sensors. Experiments under the challenging modality-agnostic settings with arbitrary-modal inputs. The results show that our method significantly outperforms existing works by a large margin (\textbf{+2.19\%} \& \textbf{+7.25\%} on MUSES and DELIVER datasets).

This work builds upon our ECCV 2024 publication~\cite{zheng2025centering}, presenting significant methodological and experimental advancements in the following key aspects:

\begin{itemize}
    \item \textbf{(I)} We enhance the multi-modal aggregation module by upgrading it to the Multi-modal Interaction Module (MIM), detailed in Sec.~\ref{sec:MIM}. MIM leverages both channel-wise and spatial-wise information as guidance, enabling more effective multi-modal feature interaction and integration.
    
    \item \textbf{(II)} We expand on the concept of centering modality values by introducing a hierarchical modality selection mechanism across multi-scale feature spaces, improving the adaptability of our framework to diverse modalities.

    \item \textbf{(III)} To further advance multi-modal fusion as well as arbitrary modality fusion, we propose the Multi-scale Arbitrary-modal Selection Module (MASM), described in Sec.~\ref{Sec:MASM}. This module dynamically fuses modality-agnostic scene features at every feature granularity during training, ensuring the backbone model remains robust to arbitrary-modal inputs during inference.

    \item \textbf{(IV)} Our experimental evaluations now span both real-world and synthetic benchmarks, including DELIVER~\cite{zhang2023delivering} in Table~\ref{Tab:DELIVER} and MUSES~\cite{brodermann2025muses} in Table~\ref{Tab:MUSES}. This marks a significant extension over the previous work, which only utilized synthetic multi-sensor datasets.

    \item \textbf{(V)} We implement the MAGIC++ framework across a diverse range of segmentation backbone models, such as SegFormer~\cite{xie2021segformer}, Swin Transformer~\cite{liu2022swin}, and Pyramid Vision Transformer~\cite{wang2021pyramid}, as illustrated in Table.~\ref{Tab:MUSES}, Table~\ref{Tab:DELIVER}, and Table~\ref{Tab:swinpvt}, covering a spectrum of architectures from lightweight to high-performance, thereby demonstrating its versatility and scalability.

    \item \textbf{(VI)} Finally, we conduct extensive quantitative and qualitative analyses to ablate and validate the effectiveness of the introduced strategies and components, as depicted in Table~\ref{AB:LossFunc_DELIVER_EXTENDED}, Figure~\ref{fig:feature_vis}, and Figure~\ref{fig:TSNE}, offering deeper insights into their contributions to overall performance.
\end{itemize}

These advancements collectively elevate the capabilities of our framework, ensuring its robustness, adaptability, and effectiveness across diverse multi-modal scenarios.

\section{Related Work}
\subsection{Semantic Segmentation} 
Semantic Segmentation is a foundational task in computer vision with applications across fields like autonomous driving~\cite{zheng2023distilling,zheng2024semantics,chen2024frozen,chen2023clip,zheng2023both,zheng2023look,zhu2023good,zheng2022transformer,feng2020deep, siam2018comparative, muhammad2022vision, wang2022sfnet, li2022self, xiao2023baseg, fantauzzo2022feddrive, nesti2022evaluating, cheng2022cenet}. Traditional methods leverage either convolutional or self-attention mechanisms. Fully convolutional networks (FCNs)~\cite{long2015fully} pioneers end-to-end pixel-wise classification for segmentation, later enhanced by approaches leveraging multi-scale features~\cite{chen2017deeplab, chen2018encoder, hou2020strip, zhao2017pyramid}, attention mechanisms~\cite{choi2020cars, fu2019dual, huang2019ccnet, yuan2021ocnet}, boundary cues~\cite{borse2021inverseform, ding2019boundary, gong2021boundary, li2020improving, takikawa2019gated}, and contextual priors~\cite{hu2019acnet, lin2017refinenet, yu2020context, zhang2018context}. More recently, transformers have been explored for segmentation, showing promise in handling long-range dependencies~\cite{strudel2021segmenter, zheng2021rethinking, xie2021segformer, gu2022multi, zhang2022topformer, zhu2021unified, wang2022rtformer, xu2022multi, zhang2022segvit, liu2021swin, liu2022swin, wang2021pyramid, wang2022pvt}.
Although these methods achieve impressive results under ideal conditions, they often struggle in complex lighting or adverse weather scenarios. We build on these advances by incorporating two plug-and-play modules to enhance segmentation robustness across diverse sensor modalities and allow for modality-agnostic operation.

\subsection{Multi-modal Semantic Segmentation}
Multi-modal Semantic Segmentation aims to integrate RGB with other complementary modalities such as depth~\cite{lyu2024omnibind, lyu2024unibind, lyu2024image, wang2020learning, zhou2020rgb, wang2020deep, cao2021shapeconv, chen2021spatial, ying2022uctnet, lee2022spsn, cong2022cir, ji2022dmra, wang2022learning, song2022improving}, thermal~\cite{shivakumar2020pst900, zhang2021abmdrnet, wu2022complementarity, liao2022cross, zhou2023mmsmcnet, xie2023cross, chen2022modality, pang2023caver, hui2023bridging, zhang2023efficient}, polarization~\cite{mei2022glass, xiang2021polarization}, events~\cite{alonso2019ev, zhang2021issafe, zheng2024eventdance, cao2023chasing, zhou2024exact}, and LiDAR~\cite{zhuang2021perception, yan20222dpass, wang2022multimodal, li2022deepfusion, borse2023x, zhang2023mx2m, liu2022camliflow, li2023mseg3d}. The development of advanced sensor technologies has led to significant progress in multi-modal fusion approaches, extending beyond dual-modality fusion to fully integrated multi-modal systems, such as in MCubeSNet~\cite{liang2022multimodal}, which enhances scene understanding through richer, more diverse sensor data.
From an architecture design perspective, multi-modal fusion models can be categorized into three main types: separate branches~\cite{broedermann2022hrfuser, wei2023mmanet, zhang2021abmdrnet, man2023bev}, joint branches~\cite{wang2022multimodal, chen2021spatial}, and asymmetric branches~\cite{zhang2023delivering, zhang2022cmx}. A common strategy involves treating RGB as the primary modality, with other sensors used as auxiliary inputs. For instance, CMNeXT~\cite{zhang2023delivering} adopts an RGB-centered design, utilizing other sensors as supplementary sources of information. However, RGB alone may not suffice under challenging conditions, such as at night. This limitation calls for more robust fusion models capable of leveraging the strengths of multiple modalities while minimizing reliance on any single sensor. More recently, Liu \etal \cite{liu2024fourier} have broadened the scope by establishing the concept of modality-incomplete scene segmentation, addressing both system-level and sensor-level modality deficiencies. 
% They introduced a missing-aware modal switch strategy that mitigates over-reliance on predominant mo
Differently, to address the missing modality, \aka, the modality-agnostic semantic segmentation challenge, our MAGIC++ framework employs a modality-agnostic approach, treating all input modalities equally. The model selects reliable and fragile features across hierarchical feature spaces, dynamically adapting to varying sensor conditions. This design enhances segmentation performance and ensures resilience to sensor failures, making it more robust in diverse and difficult scenarios.

\section{Methodology}
In this section, we introduce our MAGIC++ framework. As depicted in Fig.~\ref{fig:overall}, it consists of two pivotal modules: the multi-modal Interaction Module (MIM) and the Multi-scale Arbitrary-modal Selection Module (MASM). Our approach takes multiple visual modalities as inputs~\footnote{We take the modalities in DELVIER as example.}.

\subsection{Task Parametrization}

\noindent \textbf{Inputs:} Our framework processes input data from four distinct modalities~\cite{zhang2023delivering}, all captured or synthesized within the same scene. Specifically, we consider the following inputs: RGB images $\textbf{R} \in \mathbb{R}^{h \times w \times 3}$, depth maps $\textbf{D} \in \mathbb{R}^{h \times w \times C^D}$, LiDAR point clouds $\textbf{L} \in \mathbb{R}^{h \times w \times C^L}$, and event stack images $\textbf{E} \in \mathbb{R}^{h \times w \times C^E}$. Here, $C^D = C^L = C^E = 3$. In addition, the framework utilizes the corresponding ground truth labels $\textbf{Y}$, spanning $K$ categories. 
Unlike conventional approaches that process multi-modal data independently, our method handles a mini-batch $\{r, d, l, e\}$, where each sample is drawn from all modalities: $r \in \textbf{R}$, $d \in \textbf{D}$, $l \in \textbf{L}$, and $e \in \textbf{E}$. This joint processing facilitates holistic learning and enables effective fusion across modalities.

\noindent \textbf{Outputs:} Given the multi-modal data mini-batch $\{r, d, l, e\}$, the inputs are processed by the backbone network, producing multi-scale multi-modal feature representations $\{f_r^i, f_d^i, f_l^i, f_e^i\}_{i=1}^4$, as illustrated in Fig.~\ref{fig:overall}. These features are subsequently fed into the MIM and MASM modules, which fuse and select the robust and fragile features to strengthen the multi-modal features, respectively. Finally, the segmentation head utilizes the encoder's output features to produce the predictions $P_{m}$.

\subsection{MAGIC++ Architecture}
As illustrated in Fig.~\ref{fig:overall}, our MAGIC++ framework leverages state-of-the-art backbone models\footnote{We apply SegFormer~\cite{xie2021segformer}, PVTv2~\cite{wang2022pvt}, and Swin Transformer~\cite{liu2022swin} as backbones in our experiments}, such as SegFormer~\cite{xie2021segformer}, to serve as both the feature encoder and the segmentation head (SegHead) for each modality.  
The multi-modal mini-batch $\{r, d, l, e\}$ is directly processed by the encoder within the backbone, producing multi-scale, high-level feature representations $\{f_r^i, f_d^i, f_l^i, f_e^i\}_{i=1}^4$ for the respective modalities. This operation is formulated as:  
\begin{equation}
\setlength{\abovedisplayskip}{3pt}
\setlength{\belowdisplayskip}{3pt}
\{f_r^i, f_d^i, f_l^i, f_e^i\}_{i=1}^4 = F(\{r, d, l, e\}),
\end{equation}  
where $i$ corresponds to the feature level derived from the $i$-th transformer block of the encoder.

\begin{figure*}[t!]
    \centering
    \includegraphics[width=0.90\textwidth]{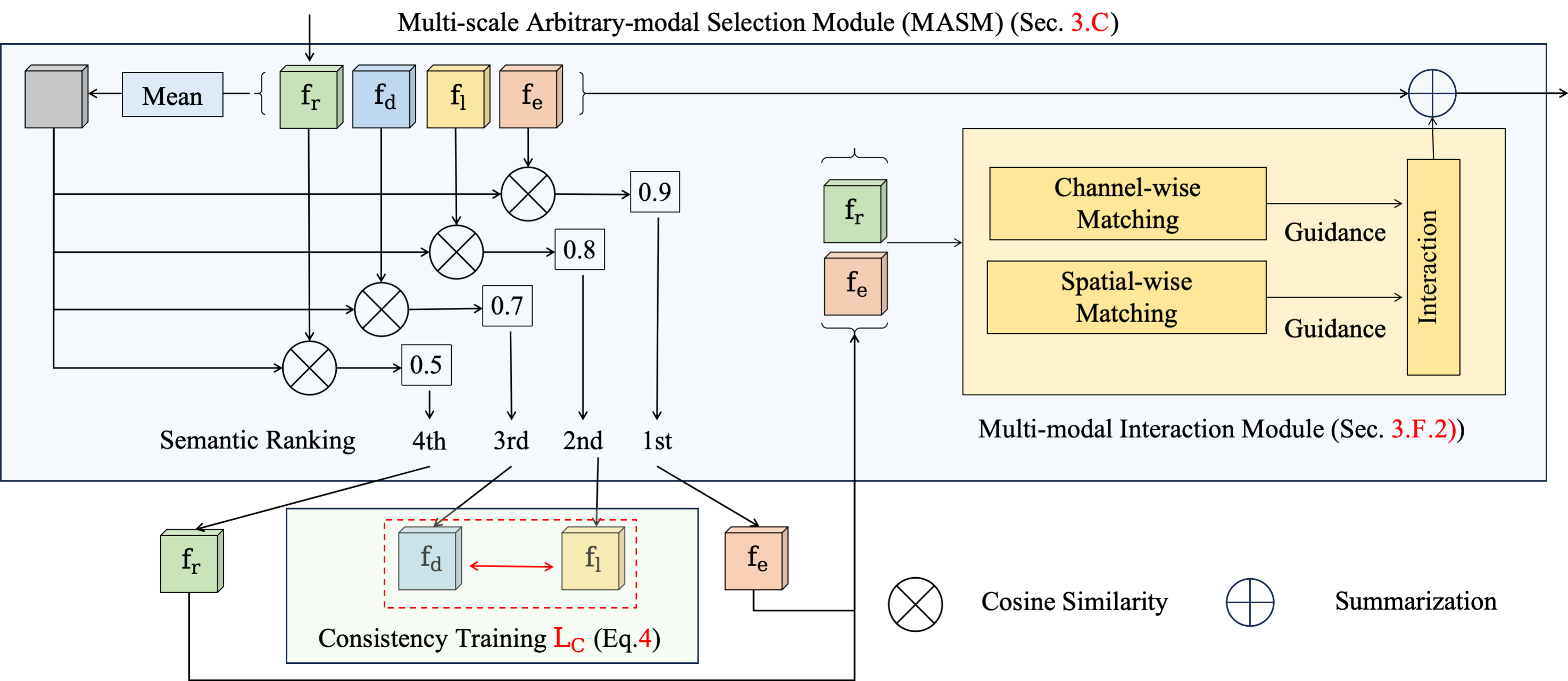}
    \caption{Illustration of the proposed plug-and-play multi-modal aggregation and multi-scale arbitrary-modal selection modules.
    }
    \label{fig:MASM}
\end{figure*}

% \begin{figure}[t!]
%     \centering
%     \includegraphics[width=0.4\textwidth]{images/AFRM.pdf}
%     \caption{Overall framework of our MAGIC++ framework, incorporates plug-and-play multi-modal aggregation and multi-scale arbitrary-modal selection modules.
%     }
%     \label{fig:ASM}
% \end{figure}

\subsection{Multi-Scale Arbitrary-Modal Selection Module}
\label{Sec:MASM}
To facilitate multi-scale feature selection, we first introduce the Multi-Scale Arbitrary-Modal Selection Module (MASM), which is employed during training to leverage the most robust features that \textit{enhance predictive accuracy} at each feature scale within the framework. The incorporation of the most fragile features—those extracted from challenging input data samples—serves to \textit{reinforce the framework's resilience against missing modalities} in such scenarios. As illustrated in Fig.~\ref{fig:overall}, our MASM consists of three primary components: cross-modal semantic similarity ranking, cross-modal semantic consistency training, and the Multi-Modal Interaction Module (MIM). We will now detail the first two components.

\subsubsection{Cross-Modal Semantic Similarity Ranking}
% Multi-modal data is characterized by significant diversity, encompassing a wide range of environmental conditions. For instance, the \textit{DELIVER} dataset~\cite{zhang2023delivering} features four distinct environmental scenarios and five episodes of partial sensor malfunctions. Such variability in real-world environments presents challenges for neural networks, making it difficult to accurately identify robust and fragile modalities at the feature level. Effectively addressing this challenge is crucial for developing a resilient multi-modal framework.

The nature of multi-modal data is inherently diverse, encompassing a wide range of conditions and challenges. A prime example is the \textit{DELIVER} dataset, as described in~\cite{zhang2023delivering}, which includes four distinct environmental scenarios and captures five episodes of partial sensor malfunctions. Beyond such specific cases, the complexities of real-world environments introduce even more heterogeneous challenges. Given this variability, it is essential for neural networks to effectively differentiate between robust and fragile modalities at the feature level.

To address this, integrating both the most robust and the most fragile modalities at the feature level can foster a more resilient multi-modal framework. In this approach, cross-modal semantic similarity ranking is used to compare multi-modal features ${f_r, f_d, f_l, f_e}$ against the semantic feature $f_{se}$ derived from the MAM, as presented in our previous work, MAGIC\cite{zheng2025centering}.
However, it is important to note that simply selecting high-level features is not always sufficient for tasks like semantic segmentation, especially when working with hierarchical backbones that incorporate pyramid features. This introduces the need for more nuanced selection strategies to ensure optimal performance across multi-modal tasks.

In the previous MAGIC framework, the Multi-Modal Aggregation Module (MAM) is designed to extract semantically rich features from high-level multi-modal inputs, thereby enhancing arbitrary-modal capabilities. Feature selection and ranking relied on semantic characteristics derived from various trainable layers, including convolutional layers, parallel pooling, and multi-layer perceptrons (MLPs). However, the use of multiple extraction layers for multi-scale feature selection is inherently computationally intensive and costly. Furthermore, the addition of these layers can lead to unreliable training outcomes during the learning process.

To overcome these limitations, the Multi-Scale Arbitrary-Modal Selection Module (MASM) employs similarity ranking to compare multi-modal features \(f_r, f_d, f_l, f_e\) against the mean feature \(f_m = Mean(f_r, f_d, f_l, f_e)\), generating ranked similarity scores. This approach allows MAGIC++ to avoid the introduction of additional trainable parameters while maintaining distinct multi-modal information during training, thereby facilitating easier convergence. The ranking process effectively identifies both the most robust and the most fragile modalities, forming a solid foundation for enhanced feature aggregation.

The ranking process is formalized as follows\footnote{We use \(i=1\) as an example, indicating that the selection occurs after the first transformer block}:
\begin{equation}
    \setlength{\abovedisplayskip}{3pt}
f_{rf}^1, f_{rm}^1 = \textit{Rank}(\textit{Cos}\{f_r^1, f_d^1, f_l^1, f_e^1\}, f_m),
\setlength{\belowdisplayskip}{3pt}
\end{equation}
where \(f_{rf}^1\) includes the top-ranked (most robust) and bottom-ranked (most fragile) features, while \(f_{rm}\) consists of the remaining features. The \(\textit{Rank}\) function sorts the features in descending order based on cosine similarity \(\textit{Cos}(\cdot)\). The identified features \(\{f_{rf}\}_{i=1}^4\) are subsequently passed to an additional Multi-Input Module (MIM), which aggregates them into a final feature $f_{mim}$.

\begin{figure*}[t!]
    \centering
    \includegraphics[width=0.99\textwidth]{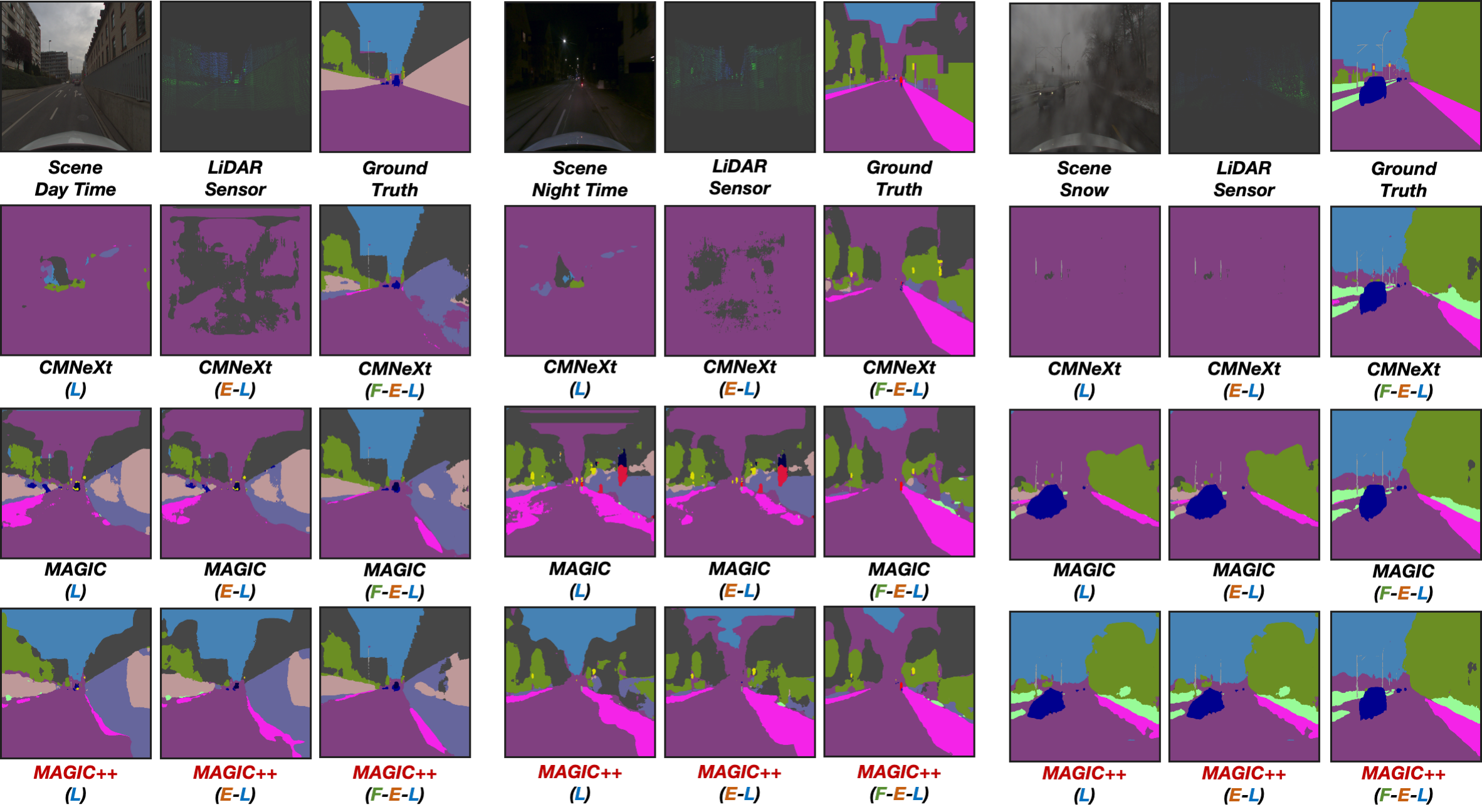}
    % \vspace{-12pt}
    \caption{Qualitative results of arbitrary inputs evaluation with CMNeXt~\cite{zhang2023delivering}, MAGIC~\cite{zheng2025centering} and the propsoed MAGIC++, using \{\textbf{F}rame, \textbf{L}iDAR, \textbf{E}vent\} on MUSES dataset~\cite{brodermann2025muses}.}
    % \vspace{-16pt}
    \label{MUSES:vis_compare}
\end{figure*}

\subsubsection{Multi-modal Interaction Module (MIM)}
\label{sec:MIM}
The Multi-modal Interaction Module (MIM) is designed to further refine and enhance the semantic richness of multi-scale, high-level multi-modal features, denoted as \(\{f_r^i, f_d^i, f_l^i, f_e^i\}_{i=1}^4\). This process is crucial for developing robust arbitrary-modal capabilities.
MIM focuses on centralizing the values from each input modality while simultaneously extracting complementary features through both channel-wise and spatial-wise feature matching. As depicted in Fig.~\ref{fig:MASM}, MIM facilitates comprehensive cross-modal calibration, thereby improving the extraction of multi-modal features.

\noindent \textit{Channel-Wise Feature Rectification:} The module processes selected fragile and robust features, \( f_r^i \) and \( f_d^i \), by embedding them along the spatial axis into attention vectors \( W_{f_r^i}^C \) and \( W_{f_d^i}^C \). Utilizing channel-wise attention techniques and applying both global max pooling and global average pooling~\cite{zhang2023cmx} on these features helps preserve crucial information.

\noindent \textit{Spatial-Wise Feature Rectification:} To complement the global calibration achieved by channel-wise rectification, spatial-wise rectification is employed to adjust local information. Fragile and robust feature maps are concatenated and embedded into spatial maps, consistent with the approach in~\cite{zhang2023cmx}.

The refined features are then integrated with multi-modal features to enhance the comprehensive scene information. Following the multi-scale selection and feature fusion processes, these aggregated features are summarized and sent to the segmentation head to produce predictions \(P_{m}\).
Ultimately, the cross-entropy is used as the supervision loss $\mathcal{L}_{M}$ :
\begin{equation}
\setlength{\abovedisplayskip}{3pt}
\setlength{\belowdisplayskip}{3pt}
    \mathcal{L}_{M} = -\sum_{0}^{K-1} \textit{Y}\cdot log(P_{m}).
\end{equation}
 
\subsubsection{Cross-modal Semantic Consistency Training.}  
Leveraging the multi-scale cross-modal semantic similarity ranking, the top-1 and last-1 ranked features are identified. Subsequently, we \textbf{\textit{impose semantic consistency}} training on the remaining features $\{f_{rm}^i\}_{i=1}^4$ across all feature scales (see Fig.~\ref{fig:MASM}). This approach is grounded in the intuition that the semantics of a scene should remain consistent across modalities, as the multi-modal data is captured under identical scenarios.

However, due to the inherent differences in data formats and sensor properties, directly aligning the remaining features $f_{rm}$ from different modalities is non-trivial. To address this, MASM employs the final feature $f_{mim}$ from MIM as a surrogate, implicitly aligning the correlation, \ie, cosine similarity, between the remaining features and the semantic feature. For clarity, we define $c_1 = \textit{Cos}(f_{rm}^1, f_{mim})$ and $c_2 = \textit{Cos}(f_{rm}^2, f_{mim})$ to represent these correlations.

The consistency training loss is then formulated as:
\begin{align}
\setlength{\abovedisplayskip}{3pt}
\setlength{\belowdisplayskip}{3pt}
\mathcal{L}_{C} = \sum_{0}^{K-1} \bigg(c_1 \log \frac{c_1}{\frac{1}{2} (c_1 + c_2)}
+ c_2 \log \frac{c_2}{\frac{1}{2} (c_1 + c_2)}\bigg).   
\end{align}

This implicit alignment ensures that features are aligned from a scene-semantic consistency perspective, facilitating robust cross-modal feature integration.

\noindent \textbf{Training}
We train our MAGIC++ by minimizing the total loss $\mathcal{L}$ -- a linear combination of the losses of $\mathcal{L}_{M}$ and $\mathcal{L}_{C}$:
\begin{equation}
\setlength{\abovedisplayskip}{3pt}
\setlength{\belowdisplayskip}{3pt}
    \mathcal{L} = \mathcal{L}_{M} + \beta \mathcal{L}_{C},
\end{equation}
where $\lambda$ and $\beta$ are hyper-parameters for trade-off. The MIM and MASM is only utilized in training while the inference is achieved by the backbone model, \ie, SegFormer~\cite{xie2021segformer}.

\begin{table*}[ht!]
\renewcommand{\tabcolsep}{9pt}
\caption{Results of MaSS validation with three modalities on real-world benchmark MUSES dataset using SegFormer-B0 as backbone model.}
\resizebox{\linewidth}{!}{
\begin{tabular}{c|c|c|ccccccc|c}
\midrule
\multirow{2}{*}{Method} & \multirow{2}{*}{Pub.} & \multirow{2}{*}{Training} & \multicolumn{7}{c}{MaSS mIoU} & \multirow{2}{*}{Mean}  \\ \cmidrule{4-10}
 &  &  & F & E & L & FE & FL & EL & FEL &  \\ \midrule
CMX~\cite{zhang2023cmx} & TITS 2023 & \multirow{3}{*}{} & 2.52 & 2.35 & 3.01 & 41.15 & 41.25 & 2.56 & 42.27 & 19.30 \\ \cmidrule{1-2} \cmidrule{4-11} 
CMNeXt~\cite{zhang2023delivering} & CVPR 2023 & \multirow{3}{*}{FEL} & 3.50 & 2.77 & 2.64 & 6.63 & 10.28 & 3.14 & 46.66 & 10.80 \\ \cmidrule{1-2} \cmidrule{4-11} 
Any2Seg~\cite{zheng2024learning} & ECCV 2024 &  & 44.40 & 3.17 & 22.33 & 44.51 & \textbf{49.96} & 22.63 & \textbf{50.00} & \underline{33.86} \\ \cmidrule{1-2} \cmidrule{4-11} 
MAGIC~\cite{zheng2025centering} & ECCV 2024 &  & 43.22 & 2.68 & 22.95 & 43.51 & \underline{49.05} & 22.98 & \underline{49.02} & 33.34 \\ \cmidrule{1-2} \cmidrule{4-11} 
\rowcolor{gray!10} MAGIC++ & - & - & \textbf{45.56} & \textbf{17.93} & \textbf{29.92} & \textbf{40.58} & 46.07 & \textbf{28.10} & 40.58 & \textbf{35.53}  \\ \midrule
\textit{w.r.t} MAGIC & - & - & +2.34 & +15.25 & +6.97 & -2.93 & -2.98 & +5.12 & -8.44 & +2.19 \\
% \midrule
% \multirow{2}{*}{Method} & \multirow{2}{*}{Pub.} & \multirow{2}{*}{Training} & \multicolumn{7}{c}{MaSS Acc} & \multirow{2}{*}{Mean}  \\ \cmidrule{4-10}
%  &  &  & F & E & L & FE & FL & EL & FEL &  \\ \midrule
% CMX~\cite{zhang2023cmx} & TITS 2023 & \multirow{3}{*}{} &  \\ \cmidrule{1-2} \cmidrule{4-11} 
% CMNeXt~\cite{zhang2023delivering} & CVPR 2023 & \multirow{3}{*}{FEL} & 5.93 & 5.36 & 5.39 & 11.16 & 15.57 & 7.76 & 56.23 & \\ \cmidrule{1-2} \cmidrule{4-11} 
% Any2Seg~\cite{zheng2024learning} & ECCV 2024 &  & 56.50 & 8.34 & 42.84 & 56.39 & 61.91 & 42.77 & 61.84 &  \\ \cmidrule{1-2} \cmidrule{4-11} 
% MAGIC~\cite{zheng2025centering} & ECCV 2024 &  & 55.76 & 6.36 & 41.29 & 54.80 & 61.41 & 39.33 & 53.12 & 44.58 \\ \cmidrule{1-2} \cmidrule{4-11} 
% \rowcolor{gray!10} MAGIC++ & - & - & 56.09 & 23.46 & 39.67 & 46.25 & 52.72 & 33.54 & 45.28 & 42.43 \\ \midrule
% \textit{w.r.t} MAGIC & - & - & +0.33 & +17.10 & -1.62 & -15.16 \\
\bottomrule
\end{tabular}}
\label{Tab:MUSES}
% \vspace{-12pt}
\end{table*}

\section{Experiments}

\subsection{Experimental Setup}

\noindent \textbf{Datasets.} We evaluate the MAGIC++ framework on both synthetic and real-world multi-sensor datasets. The MUSES dataset~\cite{brodermann2025muses} includes driving sequences from Switzerland, designed to address challenges posed by adverse visual conditions. It features multi-sensor data, including a high-resolution frame camera (F), an event camera (E), and MEMS LiDAR (L), which provide complementary modalities for enhanced annotation quality and robust multi-modal semantic segmentation. Each sequence is annotated with high-quality 2D panoptic labels, offering accurate ground truth for comprehensive benchmarking.
The DELIVER dataset~\cite{zhang2023delivering} consists of RGB (R), depth (D), LiDAR (L), and event (E) data across 25 semantic categories, recorded under various environmental conditions and including scenarios with sensor failures. This diversity allows for evaluations under challenging situations. We follow the official data processing and splitting protocols for both datasets.

\noindent \textbf{Implementation Details.} Experiments on both the MUSES and DELIVER dataset were conducted on 8 NVIDIA 3090 GPUs. The initial learning rate was set to \(6 \times 10^{-5}\), with polynomial decay (power of 0.9) over 200 epochs. A 10-epoch warm-up at 10\% of the initial learning rate was applied to stabilize training. The AdamW optimizer was used, and the effective batch size was set to 16 for both datasets. For consistency across benchmarks, input modalities were cropped to \(1024 \times 1024\) resolution.

\noindent \textbf{Experimental Settings.}  
Modality-agnostic Semantic Segmentation (MaSS): Expanding on the foundation of MAGIC, our MAGIC++ framework aims to enhance MaSS performance while maintaining balanced results across all modality combinations. To evaluate this, we test all possible input modality combinations and compute the average performance to derive the final mean result.

\begin{figure*}[t!]
    \centering
    \includegraphics[width=\textwidth]{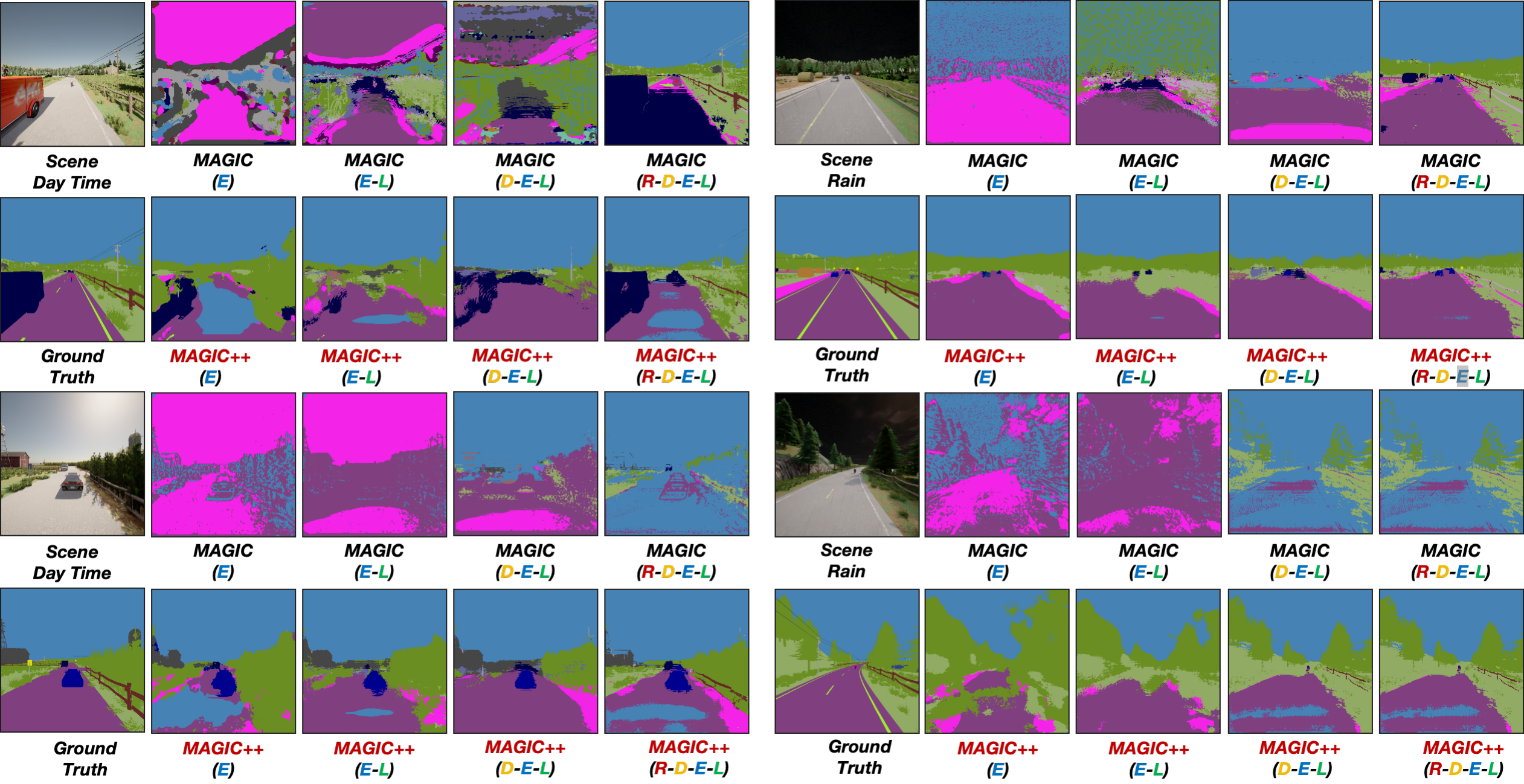}
    % \vspace{-12pt}
    \caption{Qualitative results of arbitrary inputs evaluation with CMNeXt~\cite{zhang2023delivering}, MAGIC~\cite{zheng2025centering} and the propsoed MAGIC++, using  \{\textbf{R}GB, \textbf{D}epth, \textbf{E}vent, \textbf{L}iDAR\} on DELIVER dataset~\cite{zhang2023delivering}.}
    % \vspace{-16pt}
    \label{DELIVER:vis_compare}
\end{figure*}
\begin{table*}[t!]
\renewcommand{\tabcolsep}{2.5pt}
\caption{Results of anymodal semantic segmentation validation with three modalities on synthetic benchmark DELIVER dataset using SegFormer-B0 as backbone model.}
\resizebox{\linewidth}{!}{
\begin{tabular}{c|ccccccccccccccc|c}
\midrule
\multirow{2}{*}{Method} & \multicolumn{15}{c}{MaSS mIoU} & \multirow{2}{*}{Mean}\\ \cmidrule{2-16}
 & R & D & E & L & RD & RE & RL & DE & DL & EL & RDE & RDL & REL & DEL & RDEL & \\ \midrule
CMNeXt~\cite{zhang2023delivering} & 0.86 & 0.49 & \underline{0.66} & 0.37 & 47.06 & 9.97 & 13.75 & 2.63 & 1.73 & \underline{2.85} & 59.03 & 59.18 & 14.73 & \textbf{59.18} & 39.07 & 20.77 \\ \midrule
MAGIC~\cite{zheng2025centering} & \underline{32.60} & \textbf{55.06} & 0.52 & 0.39 & \textbf{63.32} & \underline{33.02} & \underline{33.12} & \textbf{55.16} & \textbf{55.17} & 0.26 & \textbf{63.37} & \textbf{63.36} & \underline{33.32} & \underline{55.26} & \textbf{63.40} & \underline{40.49} \\ \midrule
\rowcolor{gray!10} MAGIC++ & \textbf{48.67} & \underline{52.83} & \textbf{19.03} & \textbf{18.67} & \underline{61.82} & \textbf{49.38} & \textbf{49.76} & \underline{54.39} & \underline{53.18} & \textbf{18.67} & \underline{61.76} & \underline{61.87} & \textbf{50.19} & 54.25 & \underline{61.67} & \textbf{47.74} \\ \midrule
\textit{w.r.t} MAGIC & +16.07 & -2.23 & +18.51 & +18.28 & -1.50 & +16.36 & +16.64 & -0.77 & -1.99 & +18.41 & -1.61 & -3.10 & +16.87 & -1.01 & -1.73 & +7.25 \\ 
% \midrule
% \multirow{2}{*}{Method} & \multicolumn{15}{c}{MaSS Acc} & \multirow{2}{*}{Mean}\\ \cmidrule{2-16}
%  & R & D & E & L & RD & RE & RL & DE & DL & EL & RDE & RDL & REL & DEL & RDEL & \\ \midrule
% CMNeXt~\cite{zhang2023delivering} &  \\ \midrule
% MAGIC~\cite{zheng2025centering} &  \\ \midrule
% \rowcolor{gray!10} MAGIC++ &  \\ \midrule
% \textit{w.r.t} MAGIC & \\ 
\bottomrule
\end{tabular}}
\label{Tab:DELIVER}
% \vspace{-12pt}
\end{table*}

\subsection{Experimental Results}

\subsubsection{MaSS on MUSES}

MAGIC++ demonstrates a significant advancement in handling arbitrary input modalities, overcoming challenges faced by prior methods like CMX~\cite{zhang2023cmx} and CMNeXt~\cite{zhang2023delivering}, which often struggle in scenarios involving sparse modalities such as LiDAR (L) or Event (E) data. As shown in Table~\ref{Tab:MUSES}, MAGIC++ achieves substantial improvements in MaSS on the MUSES dataset.

Compared to its predecessor MAGIC~\cite{zheng2025centering}, MAGIC++ achieves a mean performance improvement of +2.19\%, along with notable gains in specific modalities. For example, MAGIC++ outperforms MAGIC in Frame (\textbf{45.56\%} vs. 43.22\%, \textbf{+2.34\%}) and Event data (\textbf{17.93\%} vs. 2.68\%, \textbf{+15.25\%}). These results underline the ability of MAGIC++ to integrate complementary information from diverse modalities effectively.
MAGIC++ also outperforms state-of-the-art methods like Any2Seg~\cite{zheng2024learning} and CMNeXt~\cite{zhang2023delivering}. For instance, it achieves better results in LiDAR-Event (EL) combinations (\textbf{28.10\%} vs. 22.63\%, \textbf{+5.12\%}). Although Any2Seg shows a slight advantage in ELF settings, MAGIC++ demonstrates greater resilience in scenarios with limited data diversity or modality-specific noise, leveraging its multi-scale arbitrary-modal selection learning and multi-modal interaction mechanisms.
With its streamlined and modular design, MAGIC++ provides enhanced flexibility and efficiency in diverse modality combinations of the real-world sensor configurations. These results validate its effectiveness as a robust and efficient solution for arbitrary-modality segmentation, advancing multi-modal semantic segmentation capabilities.

As illustrated in Fig.~\ref{MUSES:vis_compare}, MAGIC++ demonstrates a remarkable ability to handle arbitrary combinations of input modalities, outperforming prior approaches, particularly in challenging scenarios with sparse or incomplete modalities such as LiDAR (L) or Event (E) data. Unlike CMNeXt, which struggles to maintain consistency and semantic integrity in these cases, MAGIC++ achieves robust segmentation results across diverse environmental conditions, including daytime, rain, and snow.
Compared to its predecessor MAGIC~\cite{zheng2025centering}, MAGIC++ significantly enhances performance by effectively integrating complementary information from multi-modal inputs. For instance, in the presence of only LiDAR data (L) or combined Event-LiDAR data (E-L), MAGIC++ produces more coherent and accurate segmentation maps with clearer boundaries and fewer artifacts. The improvements are particularly evident in the fused multi-modal settings (F-E-L), where MAGIC++ achieves a higher level of semantic consistency and accurately captures fine-grained details in the scene.

Furthermore, MAGIC++ demonstrates superior flexibility in adapting to different modality combinations without sacrificing performance, achieving notable gains over MAGIC and state-of-the-art methods like Any2Seg~\cite{zheng2024learning} and CMNeXt~\cite{zhang2023delivering}. Specifically, MAGIC++ outperforms Any2Seg in the LiDAR-Event (E-L) setting, with cleaner segmentation maps and better preservation of scene structure. Even in extreme cases with limited modality diversity, MAGIC++ maintains robust segmentation, thanks to its multi-scale arbitrary-modal selection (ASM) module and enhanced multi-modal interaction mechanisms.
These results underscore the ability of MAGIC++ to deliver accurate and consistent segmentation results across arbitrary modality inputs, even in complex real-world environments. With its streamlined architecture and modular design, MAGIC++ sets a new benchmark for multi-modal semantic segmentation, advancing the field by addressing the challenges posed by diverse and sparse sensor configurations.

\begin{table*}[t!]
\renewcommand{\tabcolsep}{8pt}
\caption{Results of modality-agnostic validation with three modalities.}
% \vspace{-10pt}
\resizebox{\textwidth}{!}{
\begin{tabular}{c|c|c|ccccccc|c|c}
\midrule
\multirow{2}{*}{Method} & \multirow{2}{*}{Backbone} & \multirow{2}{*}{Training} & \multicolumn{7}{c}{DELIVER dataset} & \multirow{2}{*}{Mean} & \multirow{2}{*}{$\Delta \uparrow$} \\ \cmidrule{4-10}
 &  &  & R & D & L & RD & RL & DL & RDL &  &  \\ \midrule
CMNeXt~\cite{zhang2023delivering} & Seg-B2 & \multirow{3}{*}{RDL} & 1.87 & 1.87 & 2.01 & 52.90 & 23.35 & 4.67 & \textbf{65.50} & 21.74 & - \\ \cmidrule{1-2} \cmidrule{4-12} 
MAGIC & Seg-B0 &  & 32.41 & \textbf{56.20} & 1.40 & \textbf{62.64} & 32.61 & \textbf{56.29} & \underline{62.64} & 43.46 &+21.72 \\ \cmidrule{1-2} \cmidrule{4-12} 
\rowcolor{gray!10} MAGIC++ & Seg-B0 & & \textbf{48.65} & \underline{53.61} & \textbf{8.65} & \underline{61.39} & \textbf{49.59} & \underline{53.97} & 61.80 & \textbf{48.23} & +26.49 \\ \midrule
\textit{w.r.t} MAGIC & & & +16.24 & -2.59 & +7.25 & -1.25 & +16.98 & -2.32 & -0.84 & \textbf{+4.77} & - \\ \midrule
\multicolumn{3}{c}{} & R & D & E & RD & RE & DE & RDE & Mean & $\Delta \uparrow$ \\ \midrule
CMNeXt~\cite{zhang2023delivering} & Seg-B2 & \multirow{3}{*}{RDE} & 1.75 & 1.71 & 2.06 & 53.68 & 9.66 & 2.84 & \textbf{64.44} & 19.45 & - \\ \cmidrule{1-2} \cmidrule{4-12} 
MAGIC & Seg-B0 &  & 32.96 & \textbf{55.90} & 2.15 & \textbf{62.52} & \underline{33.25} & \textbf{56.00} & \underline{62.49} & 43.61 & +24.16 \\ \cmidrule{1-2} \cmidrule{4-12} 
\rowcolor{gray!10} MAGIC++ & Seg-B0 & & \textbf{48.91} & \underline{53.26} & \textbf{11.92} & \underline{61.71} & \textbf{49.06} & \underline{52.82} & 61.83 & \textbf{48.50} & +29.05 \\ \midrule
\textit{w.r.t} MAGIC & & & +15.95 & -2.64 & +9.77 & -0.81 & +15.81 & -3.18 & -0.66 & \textbf{+4.74} & - \\ 
\bottomrule
\end{tabular}}
% \vspace{-12pt}
\label{Tab:ArbitrarySeg_3modal}
\end{table*}
\subsubsection{MaSS on DELIVER}

On the DELIVER dataset, MAGIC++ continues to demonstrate superior performance compared to its predecessor MAGIC~\cite{zheng2025centering} and other methods like CMNeXt~\cite{zhang2023delivering}, especially in scenarios involving sparse modality inputs. As detailed in Table~\ref{Tab:DELIVER}, MAGIC++ achieves consistent improvements across almost all evaluation settings.
MAGIC++ achieves an average mean performance improvement of +7.25\% over MAGIC, showcasing its ability to generalize effectively across diverse modality combinations. Notably, it excels in single-modality scenarios, achieving significant gains in RGB (\textbf{48.67\%} vs. 32.60\%, \textbf{+16.07\%}) and Event data (\textbf{19.03\%} vs. 0.52\%, \textbf{+18.51\%}). This highlights its robustness in settings where input modalities are sparse or constrained.

In pairwise modality evaluations, MAGIC++ further demonstrates its strength, achieving notable improvements in RGB-Event (RE, \textbf{49.38\%} vs. 33.02\%, \textbf{+16.36\%}) and RGB-LiDAR (RL, \textbf{49.76\%} vs. 33.12\%, \textbf{+16.64\%}). It also excels in three-modality combinations, such as REL (\textbf{50.19\%} vs. 33.32\%, \textbf{+16.87\%}), leveraging complementary information more effectively.
Compared to CMNeXt, MAGIC++ delivers significant performance boosts across most scenarios. For instance, in RGB-Event-LiDAR (REL) combinations, MAGIC++ achieves \textbf{50.19\%}, a dramatic improvement over CMNeXt's \textbf{14.73\%} (\textbf{+35.46\%}). These results highlight its ability to handle complex multi-modal data with greater precision.
By incorporating multi-scale arbitrary-modal selection learning and interaction mechanisms, MAGIC++ enables superior performance in missing modality scenarios. Its streamlined design ensures adaptability and efficiency, making it a highly robust solution for more modality scenarios involving diverse and arbitrary input modalities.

The qualitative results in Fig.~\ref{DELIVER:vis_compare} further validate the quantitative findings, showcasing how MAGIC++ produces cleaner and more semantically consistent segmentation outputs compared to MAGIC and CMNeXt. In sparse or constrained modalities like Event (E) and LiDAR (L), MAGIC++ provides accurate boundary delineations and detailed class predictions, significantly reducing artifacts and ambiguities visible in prior methods. Furthermore, in multi-modality combinations (e.g., RDEL), MAGIC++ demonstrates superior feature fusion, capturing the nuances of complex scenes effectively.

By integrating multi-scale arbitrary-modal selection learning and advanced interaction mechanisms, MAGIC++ achieves a remarkable balance of adaptability, efficiency, and performance. Its ability to handle arbitrary modality combinations ensures robust and accurate segmentation across real-world multi-modal scenarios, setting a new benchmark in multi-modal semantic segmentation.

\subsubsection{MaSS with 3 Modality on DELIVER Dataset}

Table~\ref{Tab:ArbitrarySeg_3modal} presents the results of validation on the DELIVER dataset using 3 modalities for training. The experiments evaluate the performance of CMNeXt~\cite{zhang2023delivering}, MAGIC~\cite{zheng2025centering}, and MAGIC++ across various modality combinations, highlighting the advancements brought by MAGIC++.
MAGIC++ consistently achieves the best mean performance across all configurations, demonstrating its ability to handle arbitrary modality combinations effectively. Specifically, in the RDL training setup, MAGIC++ attains a mean score of \textbf{48.23\%}, surpassing MAGIC by \textbf{+4.77\%} and CMNeXt by \textbf{+26.49\%}. The improvement is particularly evident in RGB (\textbf{48.65\%} vs. 32.41\%, \textbf{+16.24\%}) and LiDAR (\textbf{8.65\%} vs. 1.40\%, \textbf{+7.25\%}) modalities. This demonstrates the robustness of MAGIC++ in leveraging sparse and diverse modalities.

For the RDE training setup, MAGIC++ similarly outperforms MAGIC and CMNeXt, achieving a mean score of \textbf{48.50\%}, which is \textbf{+4.74\%} higher than MAGIC and \textbf{+29.05\%} higher than CMNeXt. Notably, MAGIC++ shows significant gains in Event data (\textbf{11.92\%} vs. 2.15\%, \textbf{+9.77\%}) and RGB-Event (RE) combinations (\textbf{49.06\%} vs. 33.25\%, \textbf{+15.81\%}). These results highlight the effectiveness of MAGIC++ in integrating complementary modalities and addressing the challenges posed by sparse or incomplete data.
Overall, the results indicate that MAGIC++ leverages its advanced multi-modal interaction mechanisms and multi-scale arbitrary-modal selection learning to deliver superior performance across diverse modality configurations. The substantial improvements over MAGIC and CMNeXt demonstrate its robustness and adaptability in real-world multi-modal scenarios.

\begin{table*}[t!]
\renewcommand{\tabcolsep}{8pt}
\caption{Results of modality-agnostic validation with three modalities with PVTv2 and Swin transformer as backbone and FPN as segmentation head.}
% \vspace{-10pt}
\resizebox{\textwidth}{!}{
\begin{tabular}{c|c|c|ccccccc|c|c}
\midrule
\multirow{2}{*}{Method} & \multirow{2}{*}{Backbone} & \multirow{2}{*}{Training} & \multicolumn{7}{c}{DELIVER dataset} & \multirow{2}{*}{Mean} & \multirow{2}{*}{$\Delta \uparrow$} \\ \cmidrule{4-10}
 &  &  & R & D & L & RD & RL & DL & RDL &  &  \\ \midrule
MAGIC & PVTv2-B0 &  & 36.50 & 49.43 & 7.91 & \textbf{57.56} & 35.85 & 50.55 & \textbf{57.02} & 42.12 & - \\ \cmidrule{1-2} \cmidrule{4-12} 
\rowcolor{gray!10} MAGIC++ & PVTv2-B0 & RDL & \textbf{46.74} & \textbf{50.97} & \textbf{19.36} & 55.96 & \textbf{47.19} & \textbf{51.18} & 55.68 & \textbf{46.67} & \textbf{+4.55} \\ \cmidrule{1-2} \cmidrule{4-12} 
\textit{w.r.t} MAGIC & - & & +10.24 & +1.54 & +11.45 & -1.60 & +11.34 & +0.63 & -1.34 & - & - \\ \midrule
\multicolumn{3}{c}{} & R & D & E & RD & RE & DE & RDE & Mean & $\Delta \uparrow$ \\ \midrule
MAGIC & PVTv2-B0 & & 38.85 & 49.70 & 6.44 & \textbf{58.41} & 38.42 & \textbf{52.73} & \textbf{58.10} & 43.24 & - \\ \cmidrule{1-2} \cmidrule{4-12} 
\rowcolor{gray!10} MAGIC++ & PVTv2-B0 & RDE & \textbf{47.38} & \textbf{51.43} & \textbf{19.40} & 56.79 & \textbf{47.52} & 52.30 & 56.13 & \textbf{47.28} & +4.04 \\ \cmidrule{1-2} \cmidrule{4-12} 
\textit{w.r.t} MAGIC & - & & +8.53 & +1.73 & +12.96 & -1.62 & +9.10 & -0.43 & -1.97 & - & - \\ 
\midrule
\multirow{2}{*}{Method} & \multirow{2}{*}{Backbone} & \multirow{2}{*}{Training} & \multicolumn{7}{c}{DELIVER dataset} & \multirow{2}{*}{Mean} & \multirow{2}{*}{$\Delta \uparrow$} \\ \cmidrule{4-10}
 &  &  & R & D & L & RD & RL & DL & RDL &  &  \\ \midrule
MAGIC & Swin-tiny &  & 18.21 & \textbf{45.56} & 7.48 & 52.90 & 23.85 & 47.77 & 53.55 & 35.62 \\ \cmidrule{1-2} \cmidrule{4-12} 
\rowcolor{gray!10} MAGIC++ & Swin-tiny & RDL & \textbf{37.59} & 44.11 & \textbf{14.03} & \textbf{61.39} & \textbf{49.59} & \textbf{53.97} & \textbf{61.80} & \textbf{46.07} & \textbf{+10.45} \\ \cmidrule{1-2} \cmidrule{4-12} 
\textit{w.r.t} MAGIC & - & & +19.38 & -1.45 & +6.55 & +8.49 & +25.74 & +6.20 & +8.25 & - & - \\ \midrule
\multicolumn{3}{c}{} & R & D & E & RD & RE & DE & RDE & Mean & $\Delta \uparrow$ \\ \midrule
MAGIC & Swin-tiny & & 34.12 & 1.81 & 8.70 & 28.79 & \textbf{41.59} & 4.99 & 34.50 & 22.07 & - \\ \cmidrule{1-2} \cmidrule{4-12} 
\rowcolor{gray!10} MAGIC++ & Swin-tiny & RDE & \textbf{38.82} & \textbf{45.26} & \textbf{12.33} & \textbf{54.18} & 38.96 & \textbf{47.53} & \textbf{53.48} & \textbf{41.51} & \textbf{+19.44} \\ \cmidrule{1-2} \cmidrule{4-12} 
\textit{w.r.t} MAGIC & - & & +4.70 & +43.45 & +3.63 & +25.39 & -2.63 & +42.54 & +18.98 & - & - \\ 
\bottomrule
\end{tabular}}
% \vspace{-12pt}
\label{Tab:swinpvt}
\end{table*}
\subsubsection{MaSS with PVTv2 and Swin on the DELIVER Dataset}
The MAGIC++ framework is designed with plug-and-play modularity, allowing it to pair seamlessly with various segmentation backbones featuring hierarchical feature extraction, such as PVTv2~\cite{wang2022pvt} and Swin Transformer~\cite{liu2022swin}. Table~\ref{Tab:swinpvt} presents the results of MaSS validation on the DELIVER dataset, comparing MAGIC and MAGIC++ under two training configurations: RDL and RDE. The results highlight the substantial improvements MAGIC++ achieves over MAGIC across different backbone models and modality combinations.
For the \textbf{PVTv2-B0 backbone}, MAGIC++ consistently outperforms MAGIC in both RDL and RDE training setups. Under the RDL configuration, MAGIC++ achieves a mean score of \textbf{46.67\%}, representing an improvement of \textbf{+4.55\%} over MAGIC. Notable performance gains are observed in individual modalities, including RGB (\textbf{46.74\%} vs. 36.50\%, \textbf{+10.24\%}) and LiDAR (\textbf{19.36\%} vs. 7.91\%, \textbf{+11.45\%}). Similarly, in the RDE configuration, MAGIC++ achieves a mean score of \textbf{47.28\%}, surpassing MAGIC by \textbf{+4.04\%}. Significant improvements are seen in Event (\textbf{19.40\%} vs. 6.44\%, \textbf{+12.96\%}) and RGB (\textbf{47.38\%} vs. 38.85\%, \textbf{+8.53\%}). These results emphasize MAGIC++'s robustness in effectively integrating diverse modalities when paired with the PVTv2-B0 backbone.

For the \textbf{Swin-tiny backbone}, MAGIC++ delivers even more pronounced performance gains over MAGIC. In the RDL configuration, MAGIC++ achieves a mean score of \textbf{46.07\%}, improving by \textbf{+10.45\%} over MAGIC. This improvement is particularly evident in RGB (\textbf{37.59\%} vs. 18.21\%, \textbf{+19.38\%}) and LiDAR (\textbf{14.03\%} vs. 7.48\%, \textbf{+6.55\%}). Under the RDE configuration, MAGIC++ attains a mean score of \textbf{41.51\%}, representing a significant gain of \textbf{+19.44\%} over MAGIC. The largest improvements are observed in Depth (\textbf{45.26\%} vs. 1.81\%, \textbf{+43.45\%}) and RGB (\textbf{38.82\%} vs. 34.12\%, \textbf{+4.70\%}). These results underscore MAGIC++'s ability to leverage Swin-tiny's capabilities for multi-modal segmentation effectively.
In summary, MAGIC++ demonstrates consistent and substantial improvements over MAGIC across both backbones and training configurations. The gains in individual modalities and mean performance underscore the effectiveness of MAGIC++'s advanced multi-modal interaction mechanisms and arbitrary-modal selection design.

% \begin{table*}[t!]
% % \renewcommand{\tabcolsep}{6pt}
% \caption{Results of anymodal semantic segmentation validation with three modalities on material segmentation benchmark MCubeS dataset using SegFormer-B0 as backbone model.}
% \resizebox{\linewidth}{!}{
% \begin{tabular}{c|ccccccccccccccc|c}
% \midrule
% \multirow{2}{*}{Method} & \multicolumn{15}{c}{Modality-agnostic Semantic Segmentation} & \multirow{2}{*}{Mean}\\ \cmidrule{2-16}
%  & I & N & D & A & IN & ID & IA & ND & NA & DA & IND & INA & IDA & NDA & INDA & \\ \midrule
% CMNeXt~\cite{zhang2023delivering} & 0.90 & 0.88 & 0.54 & 0.72 & 25.58 & 26.61 & 25.61 & 2.67 & 1.55 & 1.94 & 25.46 & 33.51 & 34.24 & 3.61 & 35.60 \\ \midrule
% MAGIC~\cite{zheng2025centering} & 40.00 & 34.46 & 27.34 & 25.30 & 42.11 & 41.57 & 41.63 & 37.55 & 35.44 & 28.64 & 45.47 & 40.34 & 43.58 & 36.77 & 42.57 &  \\ \midrule
% MAGIC++ & 38.64 & 31.50 & 25.86 & 21.34 & 38.54 & 38.38 & 39.05 & 32.34 & 32.50 & 27.96 & 38.50 & 38.77 & 38.38 & 32.73 & 38.56 \\ \midrule
% \textit{w.r.t} MAGIC &  \\ 
% \bottomrule
% \end{tabular}}
% \label{Tab:DELIVER}
% \vspace{-12pt}
% \end{table*}

\begin{figure*}[t!]
    \centering
    \includegraphics[width=0.99\textwidth]{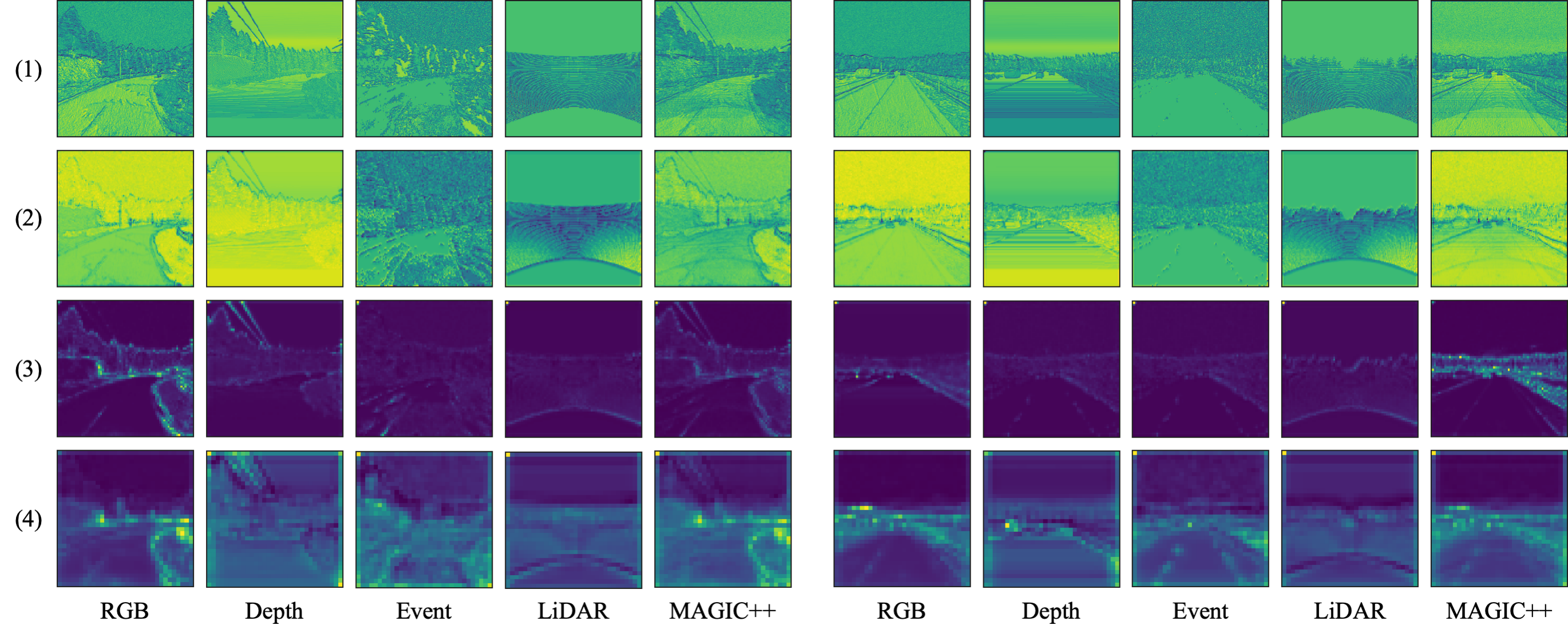}
    % \vspace{-12pt}
    \caption{Visualization of multi-scale multi-modal features and the fused MAGIC++ features. The scales correspond to: (1) $\frac{H \times W}{4}$, (2) $\frac{H \times W}{8}$, (3) $\frac{H \times W}{16}$, and (4) $\frac{H \times W}{32}$. Each column represents a modality: RGB, Depth, Event, LiDAR, and the fused MAGIC++ features.}
    % \vspace{-16pt}
    \label{fig:feature_vis}
\end{figure*}
\begin{table*}[t!]
\renewcommand{\tabcolsep}{4pt}
\caption{Ablation study of different loss function combinations on the DELIVER dataset.}
\resizebox{\linewidth}{!}{
\begin{tabular}{c|cc|ccccccccccccccc|c}
\toprule
\multirow{2}{*}{Backbone} & \multicolumn{2}{c|}{Loss} & \multicolumn{15}{c|}{MaSS mIoU} & \multirow{2}{*}{Mean} \\ 
\cmidrule{2-18}
 & $\mathcal{L}_M$ & $\mathcal{L}_C$ & R & D & E & L & RD & RE & RL & DE & DL & EL & RDE & RDL & REL & DEL & RDEL &  \\ 
\midrule
\multirow{2}{*}{\small{Seg-B0}} 
& \Checkmark & - & 47.80 & 51.23 & 16.95 & 14.91 & 60.84 & 48.21 & 48.66 & 52.17 & 51.27 & 22.91 & 60.61 & 60.83 & 60.83 & 48.88 & 60.46 & 47.10 \\ 
& \Checkmark & \Checkmark & 48.67 & 52.83 & 19.03 & 18.67 & 61.82 & 49.38 & 49.76 & 54.39 & 53.18 & 18.67 & 61.76 & 61.87 & 50.19 & 54.25 & 61.67 & 47.74 \\ \midrule
- & - & - & +0.87 & +1.60 & +2.08 & +3.76 & +0.98 & +1.17 & +1.10 & +2.22 & +1.91 & -4.24 & +1.15 & +1.04 & -10.64 & +5.37 & +1.12 & +0.64 \\
\bottomrule
\end{tabular}}
\label{AB:LossFunc_DELIVER_EXTENDED}
\end{table*}
\section{Ablation Study}

\subsection{Ablation Study on Loss Function Combinations.}
As shown in Table~\ref{AB:LossFunc_DELIVER_EXTENDED}, our proposed loss functions $\mathcal{L}_M$ and $\mathcal{L}_C$ contribute to consistent improvements in multi-modal semantic segmentation performance. Specifically, employing only $\mathcal{L}_M$ achieves a mean mIoU of \textbf{47.10\%}. Notably, the inclusion of the consistency loss $\mathcal{L}_C$ further enhances performance across all modalities, resulting in a mean mIoU improvement of \textbf{+0.64\%}, reaching \textbf{47.74\%}.

A closer analysis reveals that the improvements are consistent across individual modalities (R, D, E, L) as well as their combinations (e.g., RD, RDE, and RDEL). For example, the RDE combination improves from \textbf{60.61\%} to \textbf{61.76\%}, while the comprehensive RDEL setup achieves a final mIoU of \textbf{61.67\%} when both $\mathcal{L}_M$ and $\mathcal{L}_C$ are applied. These results validate the efficacy of our consistency loss $\mathcal{L}_C$ in further refining multi-modal feature alignment and boosting overall segmentation performance.

\subsection{Visualization of Multi-Scale Multi-Modal Features.}
In Fig.~\ref{fig:feature_vis}, we present a comprehensive visualization of multi-scale features across different modalities: RGB, Depth, Event, LiDAR, and the fused MAGIC++ features. The scales correspond to progressively reduced resolutions: (1) $\frac{H \times W}{4}$, (2) $\frac{H \times W}{8}$, (3) $\frac{H \times W}{16}$, and (4) $\frac{H \times W}{32}$, all the features are resized for better visualization.
Notably, the features extracted at coarser scales (e.g., $\frac{H \times W}{16}$ and $\frac{H \times W}{32}$) highlight global structural patterns across all modalities, while the finer scales (e.g., $\frac{H \times W}{4}$ and $\frac{H \times W}{8}$) retain more detailed and localized information. Importantly, the fused MAGIC++ features consistently exhibit richer and more complete semantic information compared to individual modalities. This validates the effectiveness of our multi-scale arbitrary-modal selection module in adaptively leveraging the most robust modalities at various scales to compensate the most fragile modalities, and further improves the multi-modal fusion and modality-agnostic learning ability of MAGIC++.

Furthermore, the interaction between modalities facilitated by our multi-modal interaction module ensures complementary feature learning. For instance, as shown in rows (3) and (4), MAGIC++ features effectively integrate the distinctive patterns from Event and LiDAR modalities while preserving fine-grained details from RGB and Depth inputs. These visualizations underscore the importance of multi-scale fusion and robust modality interaction, particularly in capturing complex scene representations for semantic segmentation.

\begin{figure*}[t!]
    \centering
    \includegraphics[width=0.99\textwidth]{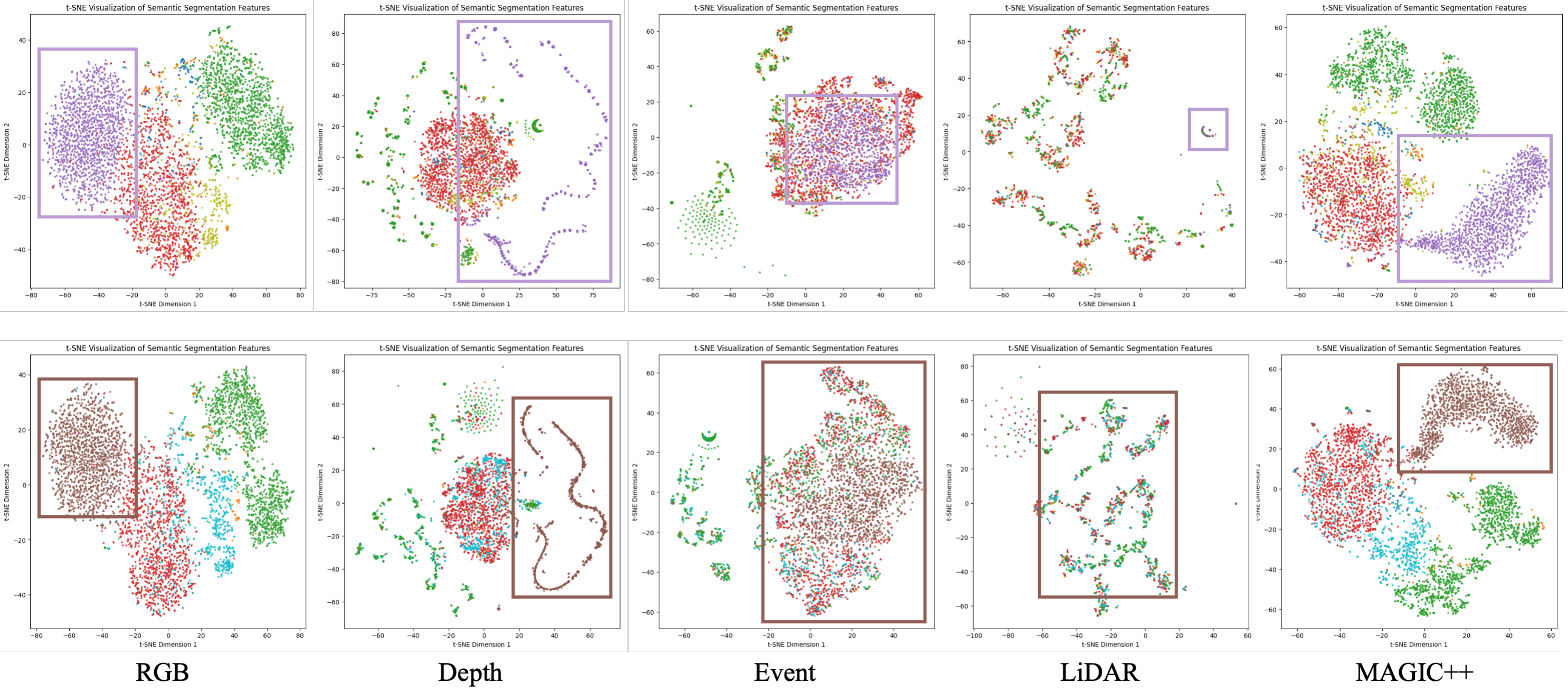}
    % \vspace{-12pt}
    \caption{t-SNE visualization of multi-modal features and the fused MAGIC++ features. Each column corresponds to a specific modality: RGB, Depth, Event, LiDAR, and the fused MAGIC++ features. The visualization demonstrates the separation and clustering of features across different scales and modalities, highlighting the effectiveness of the MAGIC++ module in integrating multi-modal information.}
    % \vspace{-16pt}
    \label{fig:TSNE}
\end{figure*}

\section{Discussion}
\subsection{Discussion on the Multi-modal Performance of MAGIC++}

The experimental results across all comparison tables (Table~\ref{Tab:MUSES}, Table~\ref{Tab:DELIVER}, Table~\ref{Tab:swinpvt}, and Table~\ref{Tab:ArbitrarySeg_3modal}) highlight the limitations of traditional multi-modal approaches, particularly when contrasted with the MaSS evaluation enabled by MAGIC++. While existing methods, such as CMNeXt~\cite{zhang2023delivering}, demonstrate competitive performance in carefully controlled multi-modal scenarios—where training configurations are manually tailored to match evaluation conditions—they struggle to maintain robustness when faced with arbitrary or sparse modality inputs. This underscores a critical weakness in handling diverse or incomplete modality combinations, which MAGIC++ effectively addresses.

It is worth noting that MAGIC++ under-performs in some controlled multi-modal evaluation settings. For instance, in the RDEL-to-RDEL scenario\footnote{Training with all four modalities and evaluation with all four modalities.} shown in Table~\ref{Tab:DELIVER}, MAGIC++ achieves \textbf{61.67\%}, slightly below MAGIC's \textbf{63.40\%}. However, when evaluated with stronger backbone models, such as the Swin Transformer~\cite{liu2022swin}, MAGIC++ consistently outperforms MAGIC in both traditional multi-modal evaluation and MaSS settings. As shown in Table~\ref{Tab:swinpvt}, MAGIC++ achieves a significant performance boost, with \textbf{53.48\%} compared to MAGIC's \textbf{34.50\%} in RDE scenarios, and \textbf{61.80\%} compared to MAGIC's \textbf{53.55\%} in RDL scenarios.
These findings reveal that while MAGIC++ may show marginally lower performance in simpler, controlled multi-modal evaluations, it excels when paired with stronger backbone models and evaluated across diverse or arbitrary modality combinations. This indicates that MAGIC++ is not only more robust in handling real-world-like scenarios but also capable of leveraging advanced backbone architectures to achieve superior performance across both traditional and MaSS evaluation settings.

\subsection{t-SNE Visualization of Multi-Modal Features.}
In Fig.~\ref{fig:TSNE}, we present t-SNE visualizations of the feature embeddings for different modalities: RGB, Depth, Event, LiDAR, and the fused MAGIC++ features. Each column corresponds to a specific modality. These visualizations demonstrate the separation and clustering of features across modalities, providing insights into the effectiveness of our proposed framework.
The individual modalities, such as RGB and Depth, show reasonable clustering for semantic classes, but significant overlaps are observed, particularly for challenging categories. In contrast, the fused MAGIC++ features exhibit more compact and well-separated clusters, underscoring the benefits of integrating multi-modal information. This is especially evident at the second line, where MAGIC++ effectively reduces intra-class variance and enhances inter-class separability compared to individual modalities.

These results highlight the ability of the MAGIC++ module to harmonize diverse multi-modal features into a unified representation. By leveraging complementary information from all modalities, the MAGIC++ module ensures robust feature learning, even under varying spatial resolutions. This demonstrates its critical role in enhancing semantic consistency and improving the overall performance of multi-modal semantic segmentation.

\section{Conclusion}

In this paper, we introduced MAGIC++, a modality-agnostic semantic segmentation framework that centers the value of every modality at every feature granularity. Addressing the challenges of robust multi-modal fusion, especially in real-world scenarios with diverse and potentially unreliable sensor inputs, MAGIC++ eliminates the traditional dependence on RGB-centric architectures. Instead, it dynamically adapts to the strengths of each modality, enhancing segmentation performance even in the presence of sensor failures or environmental noise.
Our framework comprises two key plug-and-play modules that can be integrated with various backbone models. The Multi-modal Interaction Module (MIM) efficiently processes features from input multi-modal batches, extracting complementary scene information through channel-wise and spatial-wise guidance without relying on any specific modality. Building upon MIM, the Multi-scale Arbitrary-modal Selection Module (MASM) utilizes aggregated features to rank multi-modal inputs based on similarity scores within hierarchical feature spaces. By merging both the most robust and the most fragile modalities, MASM fosters a more resilient multi-modal framework that enhances segmentation accuracy and reinforces robustness against missing or weak modalities.
Extensive experiments conducted on both real-world and synthetic benchmarks demonstrate that MAGIC++ achieves state-of-the-art performance under commonly considered multi-modal settings. Notably, in the challenging modality-agnostic setting with arbitrary-modal inputs, our method outperforms prior works by a significant margin—achieving improvements of +2.19\% on the MUSES dataset and +7.25\% on the DELIVER dataset.
This work significantly extends our previous efforts by upgrading the MIM for better feature interaction, introducing hierarchical modality selection through MASM, and validating the effectiveness of our approach with comprehensive quantitative and qualitative analyses on additional benchmarks. By fully recognizing and integrating the value of every modality at multiple feature granularities, MAGIC++ sets a new benchmark for robust and flexible multi-modal semantic segmentation.

\noindent \textbf{Future work.}
Future work may explore the integration of additional sensor modalities and further optimization of the plug-and-play modules for real-time applications. We believe that MAGIC++ paves the way toward more resilient and adaptable multi-modal perception systems, crucial for advanced robotic and autonomous systems operating in complex and dynamic environments.

% \clearpage
{
    \small
    \bibliographystyle{ieeetr}
    \bibliography{main}
}

% \clearpage
% \appendix
% \input{suppl_arxiv}

\vfill

\end{document}